# Active Phase-Encode Selection for Slice-Specific Fast MR Scanning Using a Transformer-Based Deep Reinforcement Learning Framework

Yiming Liu[1], Yanwei Pang[1], Ruiqi Jin[1], Zhenchang Wang[2]

[1]Tianjin Key Lab. of Brain Inspired Intelligence Technology, School of Electrical and Information Engineering, Tianjin University, Tianjin, China

[2]Beijing Friendship Hospital, Capital Medical University, Beijing 100050, China

Correspondence: Yanwei Pang, School of Electrical and Information Engineering, Tianjin University, Tianjin 300072, China. Email: pyw@tju.edu.cn

**Purpose:** Long scan time in phase encoding for forming complete K-space matrices is a critical drawback of MRI, making patients uncomfortable and wasting important time for diagnosing emergent diseases. This paper aims to reducing the scan time by actively and sequentially selecting partial phases in a short time so that a slice can be accurately reconstructed from the resultant slice-specific incomplete K-space matrix.

**Methods:** A transformer based deep reinforcement learning framework is proposed for actively determining a sequence of partial phases according to reconstruction-quality based Q-value (a function of reward), where the reward is the improvement degree of reconstructed image quality. The Q-value is efficiently predicted from binary phase-indicator vectors, incomplete K-space matrices and their corresponding undersampled images with a light-weight transformer so that the sequential information of phases and global relationship in images can be used. The inverse Fourier transform is employed for efficiently computing the undersampled images and hence gaining the rewards of selecting phases.

**Results:** Experimental results on the fastMRI dataset with original K-space data accessible demonstrate the efficiency and accuracy superiorities of proposed method. Compared with the state-of-the-art reinforcement learning based method proposed by Pineda et al., the proposed method is roughly 150 times faster and achieves significant improvement in reconstruction accuracy.

**Conclusions:** We have proposed a light-weight transformer based deep reinforcement learning framework for generating high-quality slice-specific trajectory consisting of a small number of phases. The proposed method, called TITLE (Transformer Involved Trajectory LEarning), has remarkable superiority in phase-encode selection efficiency and image reconstruction accuracy.

## 1 INTRODUCTION

Compared with Computed Tomography (CT), Magnetic Resonance Imaging (MRI) has advantages of zero radiation exposure and excellent soft tissue contrast, and has the disadvantage of long scan time for acquiring K-space data from which images are reconstructed. Long scan process makes patients uncomfortable, leads to motion artifacts, and wastes precious time for medical diagnosis. Therefore, reducing scan time is crucial for developing MRI devices[3,31].

Partial instead of complete scan is an important manner of reducing scan time. This paper is concentrated on 2D scan trajectory generation in Cartesian coordinate. Methods of 3D scan trajectory or 2D trajectories generations in other coordinates are also important[1, 2, 12, 33, 34] but are out of the scope of this paper.

Generally, 2D Cartesian scan is a process of forming a K-space complex matrix $\mathbf{K} \in \mathbb{C}^{F \times P}$ by frequency encoding along one axis and phase encoding along the other axis, where $F$ and $P$ stand for the number of frequencies and phases, respectively. Each element of $\mathbf{K}$ is a complex number. The matrix $\mathbf{K}$ is said to be complete if the Nyquist sampling theorem is satisfied and the image can be perfectly reconstructed from the matrix. Simply, the scan time $T$ can be expressed as $T = TR \times P$, where $TR$ stands for the repetition time. Because the scan time is mainly determined by the number of encoded phases, it is straightforward to perform fast scan by encoding and obtaining phases with $C < P$. The incomplete matrix with the $C$ selected phases is denoted by $\bar{\mathbf{K}}$. The scan acceleration factor $AF$ is defined as $AF = P / C$. There are $C_P^C$ combinations for selecting $P$ phases. Suppose $P$=368 and $C$=70, then $C_P^C \approx 10^{76}$. This paper concentrates on how to efficiently select the most important $P$ phases from such huge number of combinations so that the scan is speeded up $AF$ times and the image reconstructed from the incomplete matrix $\bar{\mathbf{K}}$ is as accurate as possible.

Denote $t_p$ as the time spent on the process of determining which phase is to be selected (encoded). It is noted that the phase selection time $t_p$ is an important factor for fast scan. As illustrated in Figure 1, the demand on small $t_p$ is strict, especially in the case of Fast Spin Echo (FSE) that multiple phases (letting the number be $m$) are encoded in one repetition time $TR$. Suppose that a phase $\theta_t$ was given after the first 180°-pulse and the resultant echo signal $O_t$ was observed at time $t$, the task of deciding an optimal phase $\theta_{t+1}$ for activating the next gradient for phase encoding at time $t_1$ has to be done during the period $t_p^*$ from the echo time $TE$ to the time $t_1$. The maximum of allowable selection time is $t_p^* = t_1 - TE < t_1 - t$. For the typical case of $TR$=2000ms, $m$=8, $TE$=40$ms$, $t$=25$ms$, $t_1$=55$ms$, and $m$=8, it holds that $t_p^*$=15$ms$. Usually, $t_p < t_p^*$ is demanded.

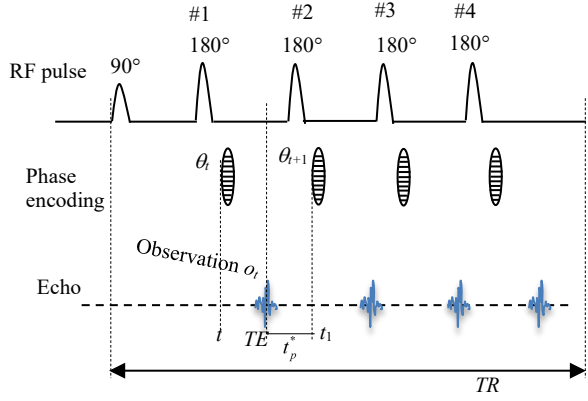

Figure 1. The Pulse-echo timing diagram and the maximum of allowable phase selection time $t_p^*$.

According to data adaptation level, the phase selection methods can be categorized into data-independent, dataset-specific, and slice-specific (subject-specific) methods. The demand on phase selection time is not problematic for the first two kinds of methods because they perform phase selection task with either no computational cost or with the offline manner. The slice-specific methods generate an optimal subset of phases specific and adaptive to the slice to be scanned and reconstructed. Therefore, this kind of methods are potentially capable of obtaining the best accuracy. However, because phase selection is performed by learning phase by phase and slice by slice, it is challenging to run efficiently. The proposed method belongs to the most challenging methods of slice-specific methods.

Data-independent methods determine a subset of phases without modelling the MRI images. A representative method is random methods which were widely used in Compressed Sensing (CS) based image reconstruction algorithms [38, 39]. In random methods, partial phases are randomly selected according to a uniform probabilistic density [38, 39] or a data-independent variable density[4,5,6, 30].

Dataset-specific methods determine a subset of phases by making use of the characteristics of the whole training data and the selected phases are fixed after training and used for any test data. Representative methods include reconstruction-error driven variable densities methods [7, 8, 9, 10], iterative low-error moving method [11], phases-embed reconstruction network [12], and Learning-based Optimization of the Under-sampling PattErn (LOUPE) method [13].

Slice-specific methods can generate a subset of phases particularly adaptive to the slice to be reconstructed [15, 16, 17, 18, 19, 20]. Representative methods include eigen-value based method [15], uncertainty-based method[17], and reinforcement learning based method [19], among which the last two methods employ deep learning and the last method outperforms the first two methods. The reinforcement learning based method proposed by Pineda et al.[19] achieved great success in slice-specific phase selection (active acquisition). Nevertheless, this method, which we call Pineda, is inefficient. Its phase selection time is 1,333$ms$ whereas the maximum of allowable selection time $t_p^*$ may be as small as 15$ms$ (Figure 1). Therefore, this method cannot be used in practical applications yet.

In this paper, we propose to overcome the inefficiency problem of the reinforcement learning based slice-specific phase selection method (i.e., the Pineda method)[19]. The novelties and contributions of the proposed method (called TITLE) are as follows.

(1) We propose a transformer-based slice-specific phase-encode selection method for determining an optimal trajectory of partial phases in a short time so that the MRI scan is accelerated with high image reconstruction accuracy. Within the framework of reinforcement learning, the decision for selecting a new phase for the next time is made according to the output (i.e., Q-value) of the transformer. Employing transformer instead of CNN (Convolutional Neural Network) is one of the novelties of the proposed compared with the Pineda method[19].

(2) The input of the deep neural network of the transformer consists of not only patches of currently reconstructed image (i.e., a state given by the environment) but also historically selected phases. This manner of forming the input to transformer makes the proposed method simultaneously utilize spatial information encoded in image and sequential information encoded in historically selected phases.

(3) We propose to generate the input image (i.e., a state given by the environment) for the transformer network by performing IFFT (Inverse Fast Fourier Transform) on the K-space data with partial phases. In the Pineda method, the input image for CNN is obtained by a time-consuming reconstruction network (i.e., a huge cascaded fully convolutional ResNet). Because the reconstruction network runs many times for producing states necessary for reinforcement learning, the Pineda method is inefficient.

(4) The employed transformer is light-weight and efficient. The light-weight transformer and efficient manner of generating input images for the transformer makes the proposed TITLE method can perform real-time slice-specific phase-encode selection in a time smaller than the maximum of allowable selection time. By contrast, the Pineda method is unable to perform phase-encode selection in a time smaller than $t_p^*$.

(5) The proposed method is not only about 150 times faster but also more accurate in reconstructing images with selected phases.

## 2 METHODS

The goal of the proposed method is to sequentially select a small number (denoted by $C$) of important phases from all candidates of phases (the number is denoted by $P$) for each slice so that the MR scan is speeded up by merely performing partial instead of full phase encoding.

### 2.1 PROBLEM FORMULATION

The learning problem of sequentially selecting partial phase can be formulated as follows.

**Training data:** Let $\mathbf{D}_K = \{\mathbf{K}_1, \mathbf{K}_2, ..., \mathbf{K}_N\}$ be a training set of $N$ full $K$-space matrices. Let $\mathbf{K}$ be a full K-space matrix sampled from $\mathbf{D}_K$. The full K-space matrix $\mathbf{K} = (\mathbf{k}_1, ..., \mathbf{k}_P) \in \mathbb{C}^{F \times P}$ consists of $F$ frequencies and $P$ phases, let each column vector correspond to a phase vector $\mathbf{k}_j \in \mathbb{C}^F$. Denote $\mathbf{I}$ as the image perfectly reconsctured from $\mathbf{K}$ by Inverse Fourior Transform (IFT) (i.e., $\mathbf{I} = IFT(\mathbf{K})$), then the training set in image domain is $\mathbf{D}_I = \{\mathbf{I}_1, \mathbf{I}_2, ..., \mathbf{I}_N\}$. The phases of $\mathbf{K}$ are descreted and then idexed by $\theta = 0, 1, ..., P-1$. Denote $\mathbf{b} \in \{0,1\}^P$ a phase-indictor vector with $b_i = 1$ indicating selection of phase $i$ and $b_i = 0$ indicating absence of phase $i$.

**Pre-selection phases:** Without loss of generality, assume there are $L$ ($L \ll P$) basic low-frequency phase vectors $\mathbf{k}_1$, $\mathbf{k}_2$, ..., and $\mathbf{k}_L$ which have been sequentially acquired at times $t+0$, $t+1$, ..., $t+L-1$ prior to selecting $M$ phases by learning. The process of sequentially selecting $C = L + M$ phases from all the $P$ phases is illustrated in Figure 2. Note that $L < M < P$ and $C = L + M < P$. The $L$ pre-selection phases, denoted by $\mathbf{A}_{pre} = \{0, 1, ..., L-1\}$, are determined without cost of computational time. Define the pre-acquired K-space matrix as $\bar{\mathbf{K}} = (\mathbf{k}_1, \mathbf{k}_2, ..., \mathbf{k}_L, 0, ..., 0) \in \mathbb{C}^{F \times P}$ where the last $P-L$ vectors are zero-valued vectors. Because $\bar{\mathbf{K}}$ contains partial phase vectors, we call it incomplete K-space matrix and denote it by $\bar{\mathbf{K}}_{pre}$. The first $L$ elements of the phase-indicator vector $\bar{\mathbf{b}} \in \{0,1\}^P$ corresponding to $\bar{\mathbf{K}}$ are identical to one and the last $P-L$ elements are equal to zero. That is, $\bar{b}_j = 1$ for $j = 0, 1, ..., L-1$ and $\bar{b}_j = 0$ for $j = L, L+1, ..., P-1$.

**Selecting the first learnable phase:** Given the $L$ pre-selection phases for phase encoding at time $t+0$, $t+1$, ..., $t+L-1$, the first task is to select a phase from the ordered set $\mathbf{A}$ of all possible candidates of phases $\mathbf{A} = \{L, L+1, ..., P-1\}$ for phase encoding at time $t+L$ according to $\bar{\mathbf{K}}$ and $\bar{\mathbf{b}}$. There are $P-L$ phases in $\mathbf{A}$. Denote the $j$-th element of $\mathbf{A}$ by $A(j)$. The problem of the task can be formualted as learning a $q$ function to estimate the importance of each candiate phase of $\mathbf{A}$:

$$\mathbf{Q} = q(\bar{\mathbf{K}}, \bar{\mathbf{b}}; t+L)) \in \mathbb{R}^{P-L} \tag{1}$$

The $q$ function in Eq.(1) outputs a $Q$ value vector $\mathbf{Q} \in \mathbb{R}^{P-L}$ with the $j$-th element $Q(j)$ of $\mathbf{Q}$ representing the importance value of phase $A(j)$. A policy function $\pi$ is then used to select a phase $j \in \mathbf{A}$ according to the $Q$ value vector $\mathbf{Q}$:

$$j = \pi(\mathbf{Q})) = \pi(Q(\bar{\mathbf{K}}, \bar{\mathbf{b}}; t+L)) . \tag{1}$$

The simplest version of $\pi$ is as follows:

$$j = \pi(\mathbf{Q})) = \arg\max_i Q(i) . \tag{2}$$

Because it is difficult to compute $Q$ value and predict $j$ directly from frequency domain data $\bar{\mathbf{K}}$, the prediction can be conducted by applying $\pi$ on the $q$ function of the image form $\bar{\mathbf{I}}$ of $\bar{\mathbf{K}}$:

$$\mathbf{Q} = q(\bar{\mathbf{I}}, \bar{\mathbf{K}}, \bar{\mathbf{b}}; t+L)) \in \mathbb{R}^{P-L}, \tag{3}$$

where $\bar{\mathbf{I}}$ is reconstructed from $\bar{\mathbf{K}}$ by a reconstructor $z$:

$$\bar{\mathbf{I}} = z(\bar{\mathbf{K}}) . \tag{4}$$

In Eq.(5), $z$ can be either the parameter-free IFT or a deep neural network (e.g., U-Net[14]). Because $\bar{\mathbf{K}}$ is an incomplete K-space matrix, we call the reconstructed image $\bar{\mathbf{I}}$ the undersampled image corresponding to $\bar{\mathbf{K}}$.

**Updation:** The phase vector acquired by encoding with the prediced phase $j$ is denoted as $\mathbf{k}_j$. Then the pre-acquried K-space matrix $\bar{\mathbf{K}}$ is updated by replacing the $j$ th column (zero vector) with $\mathbf{k}_j$. The replacement operation for $\bar{\mathbf{K}}$ is expressed as

$$\bar{\mathbf{K}} \leftarrow \bar{\mathbf{K}} \cup \mathbf{k}_j . \tag{5}$$

The image verison of $\bar{\mathbf{K}}$ is then updated by $\bar{\mathbf{I}} \leftarrow z(\bar{\mathbf{K}})$ (i.e., Eq.(5)).

Accordingly, the phase-indicator vector $\bar{\mathbf{b}}$ is updated by replacing $j$ th element (zero value) with 1 and the operation can be written as:

$$\bar{\mathbf{b}} \leftarrow \bar{\mathbf{b}} \cup 1_j . \tag{6}$$

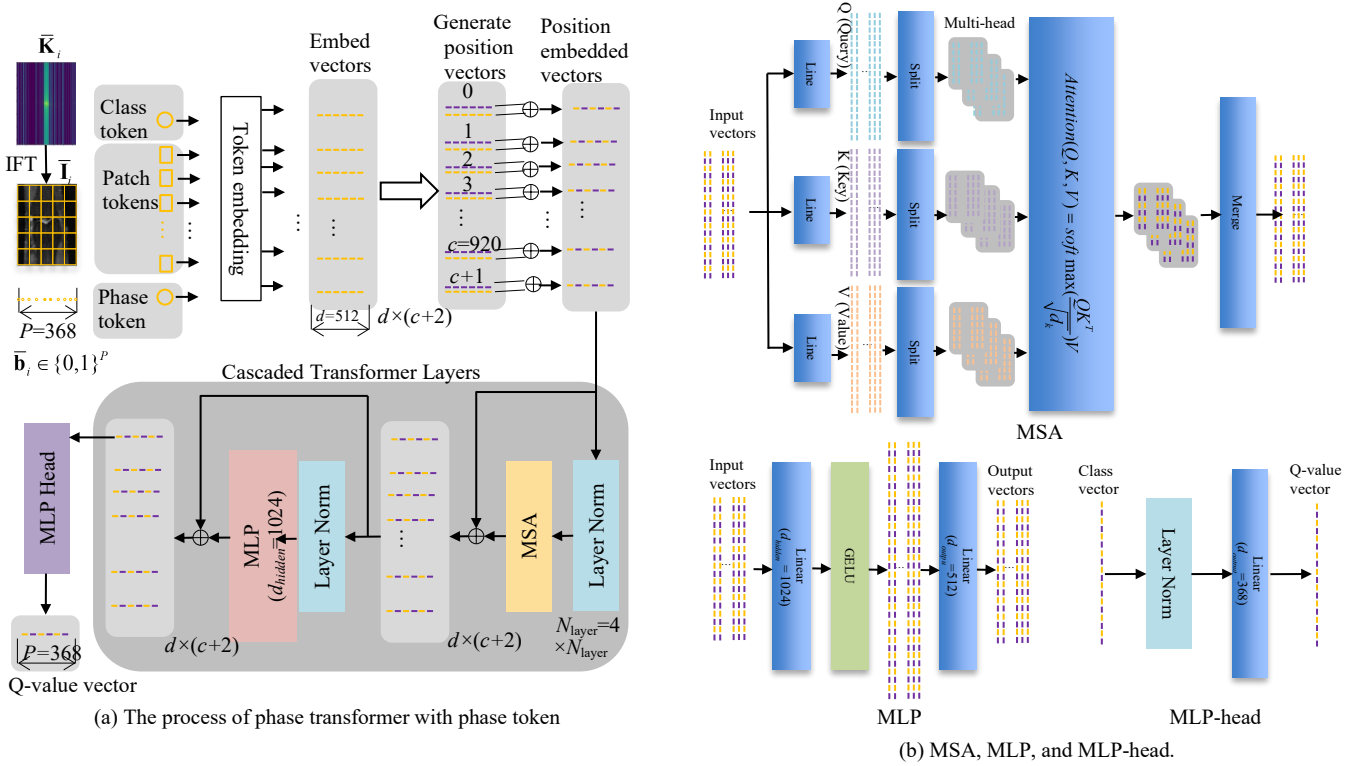

(a) The process of phase transformer with phase token

(b) MSA, MLP, and MLP-head.

Figure 2. Phase transformer. (a) The overall process (architecture) of phase transformer. (b) The details of MSA (Multihead Self Attention), MLP (Multi-Layer Perceptron), and MLP-head.

**Selecting the rest phases:** Similarly, phase selection problems for time $t+L+1$, $t+L+2$, ..., $t+L+M-1$ can be iteratively expressed as $\pi(Q(\bar{\mathbf{I}},\bar{\mathbf{K}},\bar{\mathbf{b}};t+L+1))$, $\pi(Q(\bar{\mathbf{I}},\bar{\mathbf{K}},\bar{\mathbf{b}};t+L+2))$, ..., $\pi(Q(\bar{\mathbf{I}},\bar{\mathbf{K}},\bar{\mathbf{b}};t+L+M-1))$. The iteration stops when $M$ predefined number of phases are selected and the goal of $AF = P/C$ speedup is achieved.

The objective function of the above learning problem can be written as

$$\ell = \sum_{\mathbf{K}\in\mathbf{D}_k} \sum_{s=0}^{M-1} e(Q(\bar{\mathbf{I}},\bar{\mathbf{K}},\bar{\mathbf{b}};t+L+s)\mid\mathbf{U})). \tag{7}$$

In Eq.(8), $\mathbf{U}$ stands for parameters of the model and $e(\cdot)$ is a function measuring the error of $Q$, here $Q$ is closely related to error of images reconstructed with learnt phases.

## 2.2 TRANSFORMER BASED SOLUTION

The equations in Section 2.1 are a generic form of the process of phase selection. In this subsection, we propose to implement the equations and solve the problems using vision-transformer-based reinforcement learning.

### 2.2.1 Construct the $q$ function by phase transformer *PT*

In the language of reinforcement [24, 25, 27], the $j$-th element $Q(j)$ of the output $\mathbf{Q}$ of the function $q(\bar{\mathbf{I}},\bar{\mathbf{K}},\bar{\mathbf{b}};t+L+n)$ (Eq. (4)) is termed as value of action of selecting phase $A(j)$ at time $t+L+n$ when the states are $\bar{\mathbf{I}}$, $\bar{\mathbf{K}}$, and $\bar{\mathbf{b}}$. Hereinafter, $t$ is used to stand for $t+L+n$ for the sake of notational convenience. To be consistent with terms of reinforcement learning, $Q(j)$ can be rewritten as $Q(s_t,a_t)$ with state $s_t$ and action $a_t$ at time $t$ being $s_t = \{\bar{\mathbf{I}},\bar{\mathbf{K}},\bar{\mathbf{b}}\}$ and $a_t = A(j)$, respectively. $Q(s_t,a_t)$ is defined as total reward (a.k.a., cumulative discounted reward) obtained by executing action $a_t$ in state $s_t$. The phase set $\mathbf{A} = \{L, L+1, ..., P-1\}$ defined in Section 2.1 is called action space.

We propose to compute $Q(j)$ (i.e., $Q(s_t,a_t)$ with $a_t = A(j)$ and the vector $\mathbf{Q}$ by designing a light-weight transformer (called Phase Transformer and is expressed as $PT(\cdot)$). That is, the phase transformer is used as a $q$ function:

$$\mathbf{Q} = PT(\bar{\mathbf{I}},\bar{\mathbf{K}},\bar{\mathbf{b}}|\mathbf{U}) \triangleq q(\bar{\mathbf{I}},\bar{\mathbf{K}},\bar{\mathbf{b}};t+L+n), \tag{8}$$

where $\mathbf{U}$ means the parameters of the phase transformer. The designed phase transformer has a small number of parameters and is efficient to be computed. Figure 2 shows the architecture of the proposed phase transformer which is composed of input, tokens, tokening embedding, position embedding, a cascade of transformer layers, MLP (Multi-Layer Perceptron) head, and output.

**Input of a phase transformer (*PT*):** The input (i.e., state $s_t$) of the proposed phase transformer consists of undersampled image $\bar{\mathbf{I}}$, incomplete K-space matrix $\bar{\mathbf{K}}$, and phase-indictor vector $\bar{\mathbf{b}}$, where image $\bar{\mathbf{I}}$ is obtained by applying parameter-free Inverse Fourier Transform (IFT). Different kinds of tokens are computed from the input.

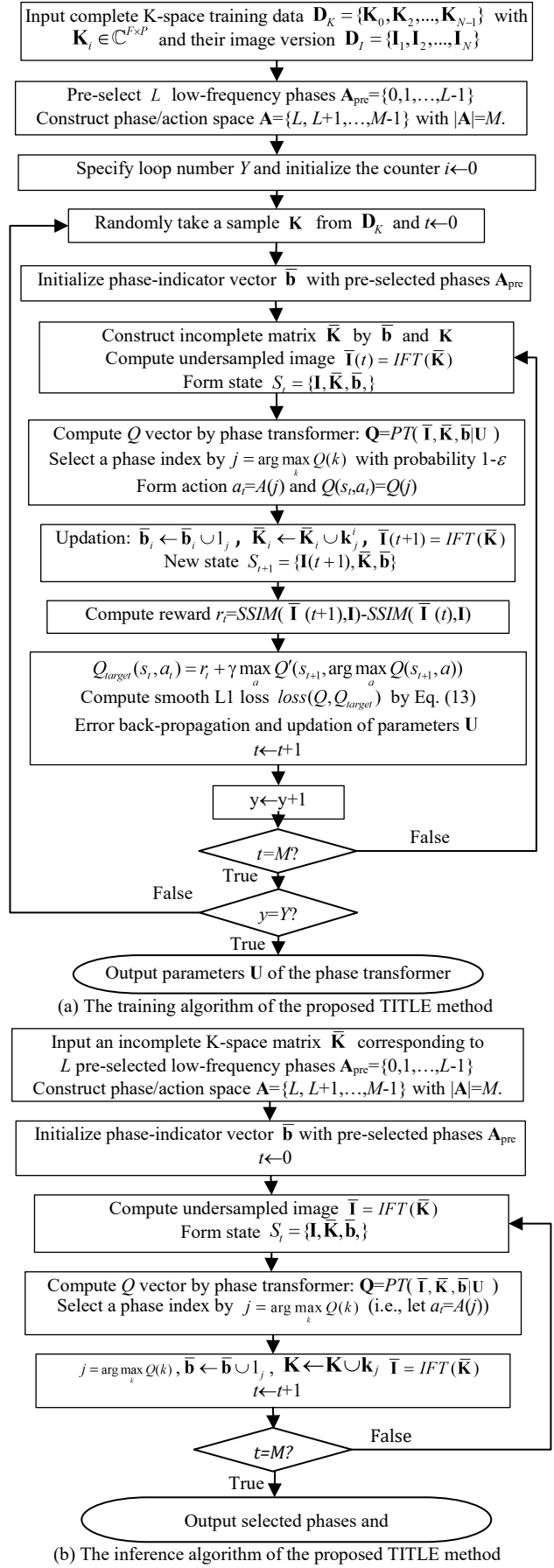

Figure 3. Training and inference flowcharts of the proposed TITLE

**Phase token:** We propose to use the phase-indictor vector $\bar{\mathbf{b}} \in \mathbb{R}^P$ as a new type of tokens called phase token. The phase token contains phases sequentially selected by now (i.e., time $t$). The historically selected phases contribute to selecting phases in the next time.

**Patch tokens and class token**: Besides, we follow the manner of classical Vision Transformer to uniformly divide the undersampling image $\bar{\mathbf{I}}$ into $c$ non-overlapping patches. A typical value of $c$ is $c = 920$. Each patch is considered as a token. Let $w$ and $h$ be the width and height of patch, respectively. The pixel number of each patch token is $w \times h$. The undersampled image $\bar{\mathbf{I}}$ and their tokens convey rich textures information which are important cues for phase selection (prediction). Class token is a learnable scalar and its initial value is randomly given. There are $c$ patch tokens, one phase token, and one class token. So there are $c + 2$ tokens in total.

**Token embedding**: All the $c + 2$ tokens are then linearly embedded (projected) onto a $d$-dimensional subspace. Each token is transformed to an embedded vector.

**Position embedding:** To preserve the spatial information of the patches and the tokens, position vectors are generated and then used for position embedding by adding the position vectors to the embedded vectors.

**Transformer layers**: The position embedded vectors form a $d \times (c+2)$ matrix. The matrix is used as the input to a cascade of transformer layers (bottom of Figure 2 (a)) to extract intrinsic features capable of predicting phases. The "$\times N_{layers}$" in bottom right of Figure 2 (a) means the number of transformer layers. From input to output, a transformer layer consists of a normalization layer called Layer Norm, an MSA (Multi-head Self Attention) layer (Figure 3 (b)) [28], another Layer Norm, and an MLP (Multi-Layer Perception) (Figure 3 (b)). The MLP has only one hidden layer. The numbers of hidden units, output units are denoted by $d_{hidden}$, and $d_{output}$. In our implementation, $d_{hidden} = 1024$, $d_{output} = 512$, and $N_{layers} = 4$ are used in our implementation for establishing a light-weight and efficient phase transformer. As shown in Figure 3 (a), there are two residual connections in a transformer layer.

**MLP-head:** The output feature matrix $\mathbf{F}$ of the last transformer layer is of size $d \times (c+2)$. The first column feature vector $\mathbf{f}_0$ of the output corresponds to class token and is taken as input to an MLP-head model for mapping $\mathbf{f}_0$ to a vector of $Q$-values, one for selecting each phase. The output process of the phase transformer is written as $\mathbf{Q} = PT(\bar{\mathbf{I}}, \bar{\mathbf{K}}, \bar{\mathbf{b}} \mid \mathbf{U})$.

**Differences from existing ViT**: Though the designed phase transformer follows the pipeline of the classical vision transformer proposed in Ref. 29 (called ViT), it is different from ViT in four aspects. (1) The proposed transformer is used for estimating Q-value (i.e., total reward obtained by selecting a phase) and the ViT is used for classification. The former is a regression problem (see Section 2.2.2) and the latter is a classification problem. (2) A new token called phase token is proposed to use in our transformer. Without the phase token, the performance of phase selection severely deteriorated (See Section 3.4). (3) The number of parameters of the proposed phase transformer is much smaller than that of ViT. In our method, the number of transformer layers (a.k.a., the number of layers of transformer decoder), the numbers of hidden units and output units of MLP are much smaller. (4) The proposed transformer can yield good results without pre-training on a large dataset whereas the performance of classical ViT heavily depends on large-scale pre-training[35].

**2.2.2 Sequential phase selection with phase-transformer-based reinforcement learning**

The $Q$-value vector output by the phase transformer $PT(\cdot \mid \mathbf{U})$ measures the total reward of selecting a phase. However, the $Q$-value vector cannot be obtained without (1) defining the reward related to image reconstruction, (2) providing supervision to the phase transformer, and (3) formulating loss function for minimizing the error of estimated $Q$-value vector. The following part is to tackle these issues.

In the context of deep reinforcement learning, the ground truth of the $Q$-value of $Q(s_t, a_t)$ can be expressed by its definition (i.e., $Q(s_t, a_t)$ is the expectation of the sum of the immediate reward $r_t$ and the discounted long-term reward $r_{t+k}$ till the destination state or equivalently the cumulative discounted reward):

$$Q(s_t, a_t) \triangleq \mathbb{E}\left[\sum_{k=0}^{M-t-k} \gamma^k r_{t+k} \mid s = s_t, a = a_t\right], \tag{9}$$

where $M$ stands for the total length of a scan trajectory (i.e., the number of phase selections (actions) executed for forming the scan trajectory) and $\gamma$ is the discount factor with $0 \leq \gamma \leq 1$.

We first state how to compute the reward $r$ and then describe how to provide ground truth of $Q$-value (i.e., target $Q$-value) in practice.

**Reward:** From Eq.(10), one can see that the supervision information for $Q(s_t, a_t)$ and $Q$-value vector can be computed if the reward is known. As in the Pineda method[19], the reward is defined as the degree of image quality improvement in image reconstruction. At time $t$, a new phase is selected according to the policy function $\pi(\mathbf{Q}) = \arg\max_j Q(j)$ (Eq.(3)). With the new phase and historically selected phases, the incomplete K-space matrix becomes $\bar{\mathbf{K}}(t)$ and the undersampled image reconstructed from $\bar{\mathbf{K}}(t)$ by $\bar{\mathbf{I}} = z(\bar{\mathbf{K}}) = IFT(\bar{\mathbf{K}})$ (i.e., Eq.(5)) becomes $\bar{\mathbf{I}}(t)$. The incomplete K-space matrix and undersampled image at time $t-1$ are denoted by $\bar{\mathbf{K}}(t)$ and $\bar{\mathbf{I}}(t)$, respectively. The ground truth of complete K-space matrix is $\mathbf{K}$ and the corresponding ground truth image (fully sampled image) is $\mathbf{I}$. Let the function $SSIM(\mathbf{x}, \mathbf{y})$ measure the Structural Similarity (see Ref. 36 for its definition and computation process) between images $\mathbf{x}$ and $\mathbf{y}$. Then the reward is

$$r_t = SSIM(\bar{\mathbf{I}}(t), \mathbf{I}) - SSIM(\bar{\mathbf{I}}(t-1), \mathbf{I}). \tag{10}$$

**Target $Q$-value and loss function**: The $Q$-value $Q(s_t, a_t)$ estimated by a deep neural network is targeted at approximating a target $Q$-value $Q_{target}(s_t, a_t)$. The $Q_{target}$ intrinsically origins from the rewards expressed in Eq.(11) and the ideal $Q_{target}$ is expressed as Eq.(10). Practically, according to the Bellman equation of optimality[37] and double DQN (Deep Q-Networks)[21,22], the $Q_{target}$ can be computed by adding the immediate reward $r_t$ with the discounted maximum of the $Q$-value of the next state $s_{t+1}$:

$$Q_{target}(s_t, a_t) = r_t + \gamma \max_a Q'(s_{t+1}, \arg\max_a Q(s_{t+1}, a)). \quad (11)$$

In Eq.(12), $Q$ and $Q'$ are obtained by the trained phase transformer and target phase transformer, respectively. Eq.(12) can be efficiently implemented by the technique of Replay Buffer (RB)[26].

The loss function $loss(Q, Q_{target})$ is a function of the difference between target $Q$-value and estimated $Q$-value. In this paper, the smooth L1 loss is employed:

$$loss(Q, Q_{target}) = \begin{cases} \frac{1}{2}(Q - Q_{target})^2 & \text{if } |Q - Q_{target}| < 1, \\ |Q - Q_{target}| - 0.5 & \text{if } |Q - Q_{target}| \geq 1. \end{cases} \quad (12)$$

It is noted that Eq.(8) is a general form of objective function and Eq.(13) is the loss function really adopted.

Once an optimal $Q$-value is obtained, a new phase can be simply selected according to the policy function $\pi(\mathbf{Q})) = \arg\max_j Q(j)$ (Eq.(3)).

### 2.2.3 Training and inference algorithms of the proposed phase selection method

Armed with the formulated problem (Section 2.1), proposed phase transformer (Section 2.2.1), and the reward and loss function (Section 2.2.2), the training algorithm for optimal parameters of phase transformer and the inference algorithm can be described as follows.

The training algorithm of the proposed TITLE method is given in Figure 3 (a). The input of the training algorithm is the complete K-space training data $\mathbf{D}_K = \{\mathbf{K}_0, \mathbf{K}_2, ..., \mathbf{K}_{N-1}\}$ and their image version $\mathbf{D}_I = \{\mathbf{I}_1, \mathbf{I}_2, ..., \mathbf{I}_N\}$. The output is the $\mathbf{U}$, the parameters of the phase transformer $PT(\cdot|\mathbf{U})$. The algorithm has an outerloop and an innter loop. The outpterloop iterates $Y$ times and a full K-space matrix is randomly selected in each iteration. The innerloop iterates $M$ ($M$ equals to the number of phases to be selected) times. In the inner loop, the following operations are conducted in order: $Q$-value is obtained by the proposed phase transformer $PT(\cdot|\mathbf{U})$, a new phase $A(j)$ is selected according to the $Q$-value, the incomplete K-space matrix $\bar{\mathbf{K}}$ is updated accroding to the selected phases, the reward of the selecting the phase $A(j)$ is computed from the image reconstructed from the updated $\bar{\mathbf{K}}$, the target $Q$-value is calculated, the loss funtion is formed with the $Q$-values, and finnaly the paramters $\mathbf{U}$ of the phase transforer are updated by error back-propogation.

The inference algorithm is given in Figure 3 (b). The input is an incomplete K-space $\bar{\mathbf{K}}_{pre}$ corresponding to the pre-selected phases. The output is the incomplete K-space matrix $\bar{\mathbf{K}}$ corresponding to the selected phases.

### 2.2.4 Image reconstruction with the selected phases

As can be seen in Section 2.2.3, the phase selection algorithm outputs the incomplete K-space matrix $\bar{\mathbf{K}}$ corresponding to the selected phases. The remained quesiton is how to reconstruct a high-quality image from $\bar{\mathbf{K}}$ [3]. In this paper, the classical U-Net[14] is trained and is used to reconstucte image from the undersampled image $\bar{\mathbf{I}}$ obained by $\bar{\mathbf{I}} = IFT(\bar{\mathbf{K}})$.

## 2.3 RELATIONSHIP TO THE PINEDA METHOD

It is noted that both the proposed method and the Pineda method[19] can be classified as deep reinforcement learning based methods. To estimate $Q$-values, a phase transformer is used in our method whereas a CNN is used in the Pineda method. The input to the phase transformer of our method includes not only images but also phase-indictor vectors conveying historical sequential information whereas the input to the CNN of the Pineda method is only images. The image input to the transformer of our method is obtained by efficient Inverse Fast Fourier Transform (IFFT) whereas the image input to the CNN of the Pineda method is obtained a time-consuming cascaded fully convolutional ResNet.

## 3 RESULTS

## 3.1 EXPERIMENTAL SETUP

The proposed TITLE method is evaluated on the challenging public fastMRI dataset where the K-space matrices were acquired by 1.5T or 3.0T scanners[23]. The K-space matrices of the single-coil knee data are of different sizes. K-space matrices with the same size are chosen. The training set, validation set, and test set are the same as that employed in the paper of the advanced Pineda method [19]. Specifically, the training, validation, and test sets consist of 19,878 K-space matrices of 536 volumes, 1,785 K-space matrices of 48 volumes, and 1,785 K-space matrices of 49 volumes, respectively. The experimental results are reported on the test set. Every complete K-space matrix is of size $F \times P = 640 \times 368$. Among the $P = 368$ phases, $L = 30$ are used as pre-selection phases. The goal is to select $M = 70$ phases from the remained $P - L = 368 - 30 = 338$ phases.

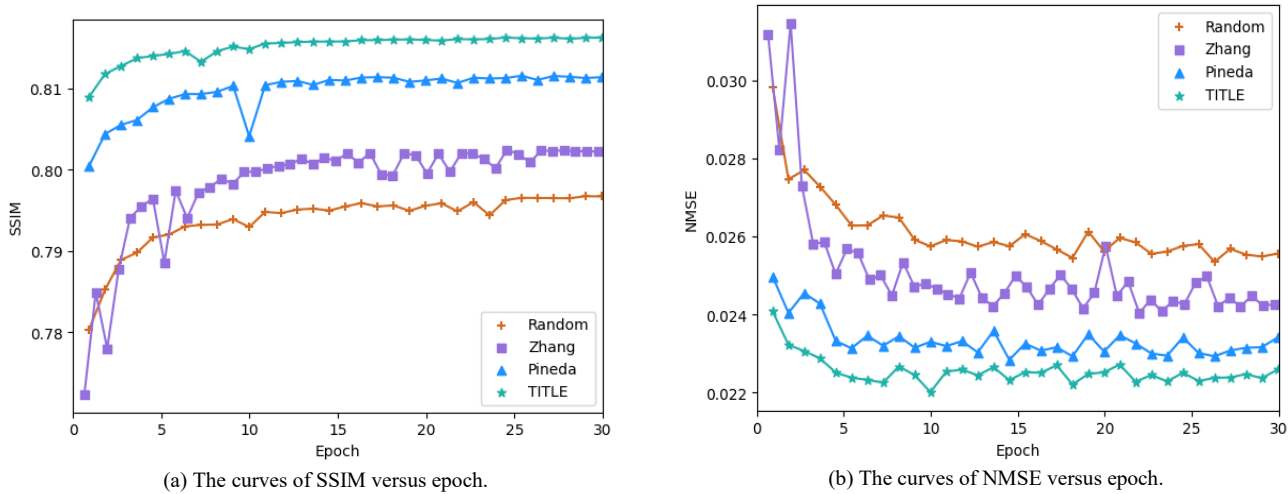

(a) The curves of SSIM versus epoch.

(b) The curves of NMSE versus epoch.

Figure 4. The curves of performance versus epoch (iteration number) of training a UNet on the results of different phase selection methods.

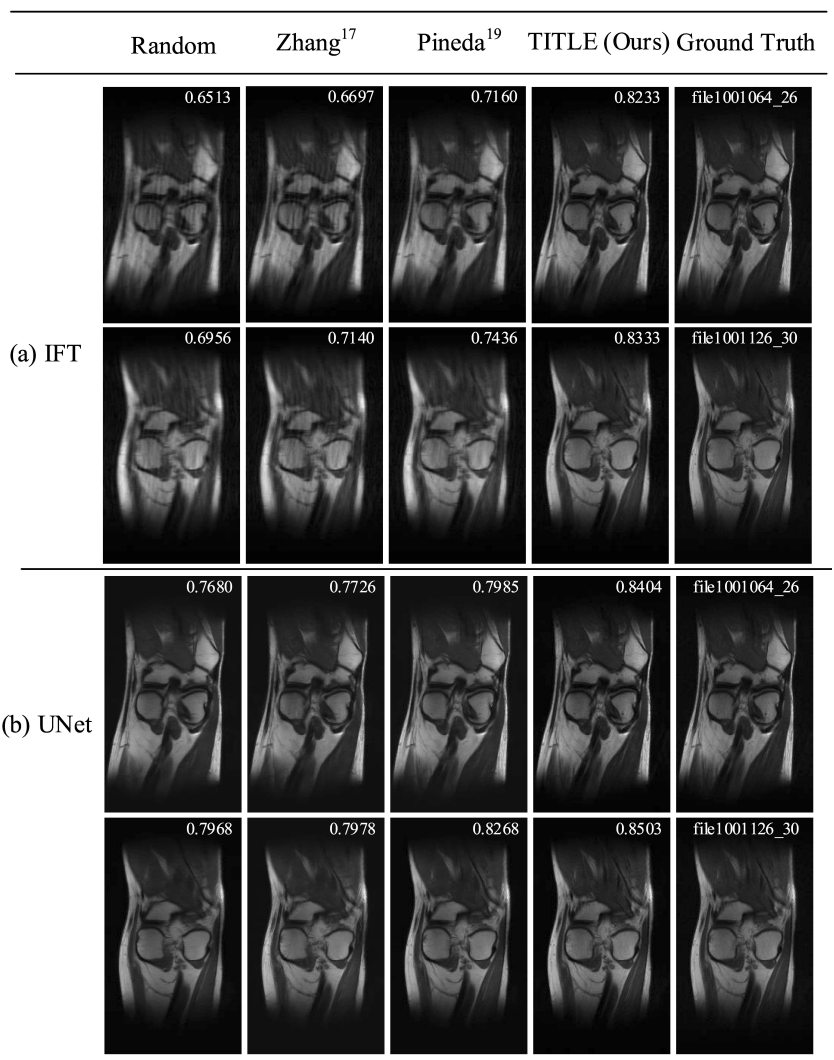

Figure 5. Examples of reconstructed images. (a) Reconstructed by IFT. (b) Reconstructed by UNet.

TABLE 1 Comparison on phase selection time (*ms*), parameter number (*M*), and GFLOPS

| Method | Random | Zhang[17] | Pineda[19] | our TITLE |
|---|---|---|---|---|
| $t_p$ | 0 | 1363 | 1333 | 9 |
| #param. | 0 | 302.71 | 302.69 | 9.52 |
| GFLOPS | 0 | 5686.53 | 5686.25 | 11.48 |

TABLE 2 Comparison on SSIM, PSNR (dB), and NSME when reconstruction on the selected phases is conducted by IFT and UNet, respectively.

| Reconstruction | Method | Random | Zhang[17] | Pineda[19] | our TITLE |
|---|---|---|---|---|---|
| IFT | SSIM | 0.6203 | 0.6315 | 0.6411 | 0.6745 |
| | PSNR | 27.10 | 27.55 | 27.76 | 29.33 |
| | NSME | 0.05276 | 0.05042 | 0.04938 | 0.04304 |
| UNet | SSIM | 0.7962 | 0.8020 | 0.8113 | 0.8163 |
| | PSNR | 33.42 | 33.66 | 34.13 | 34.36 |
| | NSME | 0.02576 | 0.02416 | 0.02341 | 0.02251 |

The proposed method is compared with three representative methods. (1) The first is called random method where 30 pre-selection phases are used and the other phases are randomly selected from the uniform distribution. This method has been widely used in CS based image reconstruction. (2) The second is called Zhang method proposed by Zhang et al. [17] where phases are selected based on a reconstruction uncertainty map. This method is the first active slice-specific phase selection method that adopts deep learning. (3) The last method, called Pineda[19], is the most advanced one, which performs sequential phase selection by integrating reinforcement learning and deep CNN. The slice-specific version of Pineda method is compared with our method.

Phase selection time $t_p$, obtained on a single RTX 3090 GPU, is used for comparison in efficiency. Averaged Structural Similarity (SSIM), Peak Signal to Noise Ratio (PSNR), and Normalized Mean Squared Error (NMSE) are adopted to measure the quality of images reconstructed with the selected phases. The architecture and hyper-parameters of the UNet for reconstructing images from selected phases is the same as in Ref. 23.

Most of the hyper-parameters of the proposed TITLE method are the same as those of the Pineda method, and the parameters mentioned above are set as: The loop number $Y$ is $10^7$, the discount factor $\gamma$ is 0.5, the size of replay buffer is 20,000, and the learning rate for updating parameters of $Q$-value network is $10^{-4}$.

## 3.2 PHASE SELECTION TIME

Table 1 gives the phase selection times ($t_p$) of different methods. Because the Random method uses a random seed to determine a phase, its phase selection time is negligible and written as zero. As can be seen from Table 1, the phase selection times of the Zhang, Pineda, and the proposed TITLE methods are 1363 *ms*, 1333 *ms*, and 9 *ms*, respectively. The proposed method takes the least time for phase selection and is roughly 150 times efficient than the Zhang and Pineda methods. A typical maximum allowable time of phase selection $t_p^*$ is 15 *ms* (Section 1). Therefore, among the learning-based methods, the proposed method is the only method satisfying $t_p < t_p^*$ and can be used in a real-time MR scan system.

The efficiency of the proposed TITLE method can also be measured in terms of parameter number and GFLOPS (Giga Floating-point Operations Per Second). It is observed from Table 1 that the parameter numbers of the Zhang method, Pineda method, and our TITLE method are 302.71M, 302.69*M*, and 9.52M, respectively. In addition, the GFLOPS of the Zhang method, Pineda method, and our TITLE method are 5686.53, 5686.25, and 11.48, respectively.

## 3.3 RESULTS OF PHASE SELECTION WITH IFT RECONSTRUCTION

The top part of Table 2 gives the performances of different methods where the images are reconstructed from the selected phases (corresponding incomplete K-space matrices) by learning-free IFT. The averaged SSIM of Random, Zhang, Pineda, and proposed TITLE methods is 0.6203, 0.6315, 0.6411, and 0.6745, respectively. Figure 5 (a) shows several examples of reconstructed images. The proposed method achieves the largest SSIM. SSIM is the most important criterion because of its consistence with eyes of human being. The averaged PSNR of Random, Zhang, Pineda, and TITLE methods is 27.10, 27.55, 27.76, and 29.33, respectively. The averaged NSME of Random, Zhang, Pineda, and proposed TITLE methods is 0.05276, 0.05042, 0.04938, and 0.04304, respectively. It is concluded that the proposed method has remarkable superiority in terms of SSIM, PSNR, and NSME.

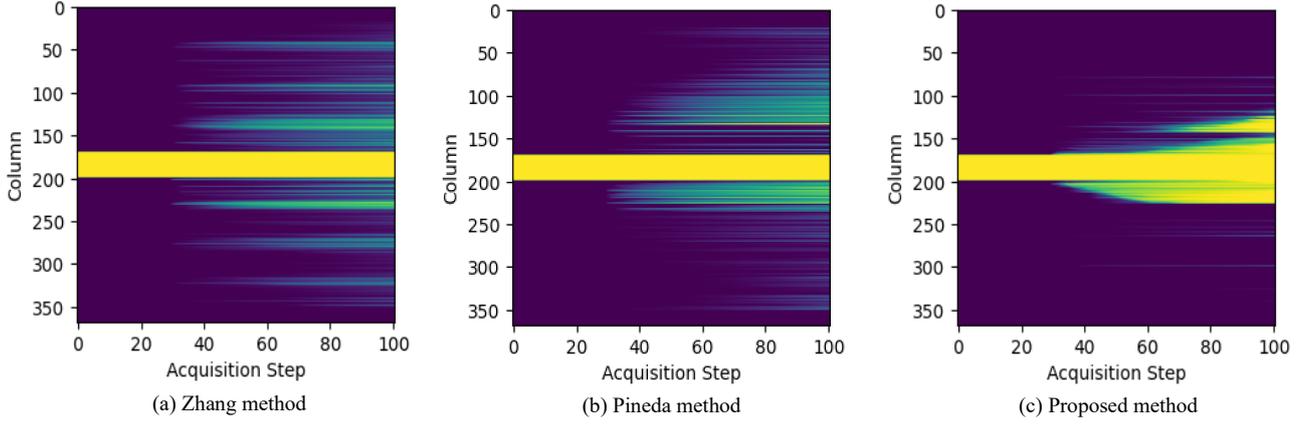

(a) Zhang method      (b) Pineda method      (c) Proposed method

Figure 6. Visualization of the average phases sequentially selected by (a) Zhang method, (b) Pineda method, and (c) the proposed TITLE method.

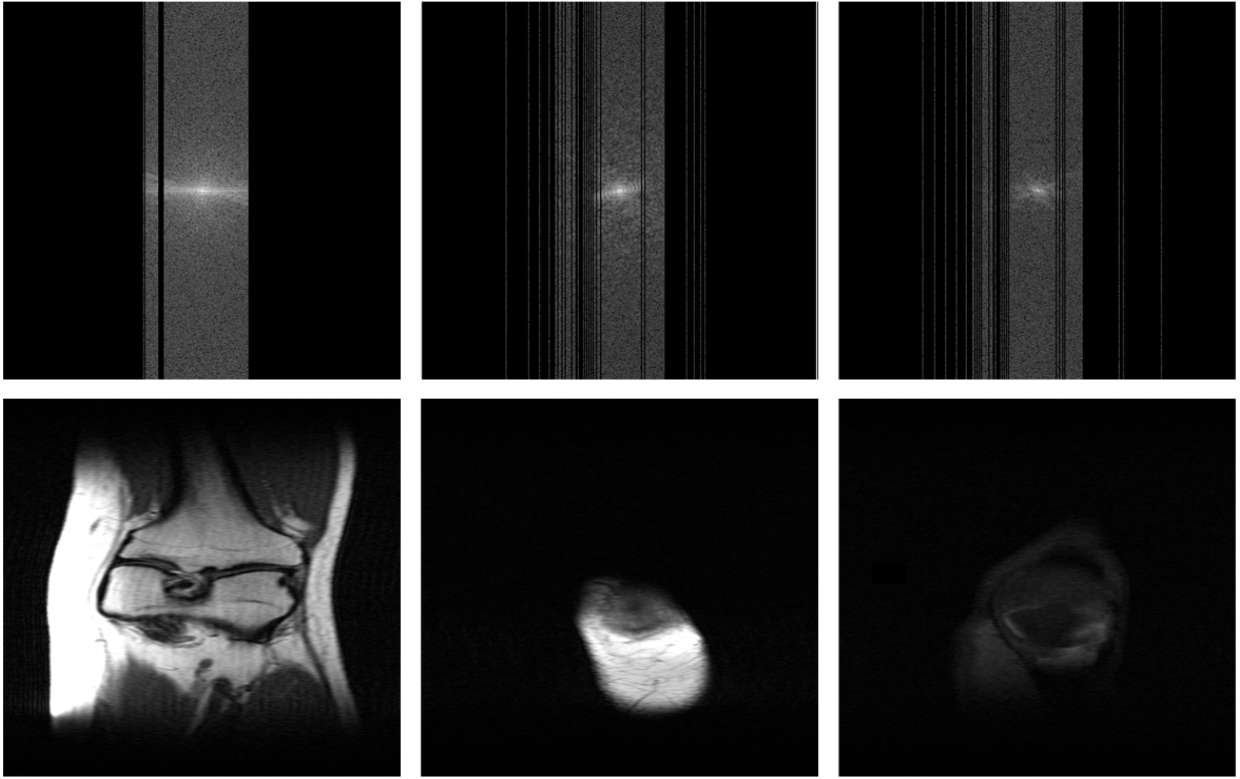

Figure 7. The phases selected by the TITLE method are adaptive to different slices. Top: The selected phase vectors (i.e., incomplete K-space matrix $\bar{\mathbf{K}}$). Bottom: The ground truth images reconstructed from complete K-space matrices.

## 3.4 RESULTS OF PHASE SELECTION WITH UNET RECONSTRUCTION

Figure 4 (a) shows how SSIM varies with epoch of UNet trained with results of phases selection methods. The SSIM, PSNR, NMSE when the UNets converge are given in the bottom part of Table 2. Specifically, the SSIM of the Radom, Zhang, Pineda, and the proposed methods are 0.7962, 0.8020, 0.8113, and 0.8163, respectively. The proposed method achieves the largest SSIM. Figure 5 (b) shows several examples of reconstructed images.

Figure 4 (b) shows the curves of NMSE versus training epoch. The NMSE of the Radom, Zhang, Pineda, and the proposed TITLE method is 0.02576，0.02416, 0.02341, and 0.02251, respectively. The proposed method is the only method whose NSME is below 0.023.

The PSNR of the Radom, Zhang, Pineda, and the proposed TITLE methods is 33.42, 33.66, 34.13, and 34.36, respectively. Table 2 shows that the proposed method is the best method in the sense of SSIM, PSNR and NSME.

TABLE 3 Comparison of the original TITLE method and its simplified version TITLE⁻ where phase token is not employed.

| Method | SSIM | PSNR | NSME |
| --- | --- | --- | --- |
| TITLE | 0.8163 | 34.36 | 0.02251 |
| TITLE⁻ | 0.8150 | 34.32 | 0.02266 |

Finally, we investigate the role of the phase tokens. The method without phase tokens (i.e., removing phase token from Figure 2 (a)) is denoted by TITLE⁻. Table 3 shows that the SSIM of TITLE and TITLE⁻ is respectively 0.8163 and 0.8150, implying removing phase token results in lower SSIM. The PSNR of TITLE and TITLE⁻ is 34.36 and 34.32, respectively. The NSME of TITLE and TITLE⁻ is 0.02251 and 0.02266, respectively. Therefore, removing phase token consistently degrades SSIM, PSNR, and NSME.

### 3.5 VISUALIZATION OF SELECTED PHASES

Visualization of the averaged phases sequentially selected by the Zhang method, Pineda method, and the proposed TITLE method are given in Figure 6.

All the methods sequentially select 70 phases besides 30 pre-selected phases. As the time $t$ grows, more and more phases are selected until 70 phases are obtained. The following phenomenon can be observed from Figure 6. (1) Averagely, compared with the Zhang method and the Pineda method, the TITLE method is concentrated on phases with low frequencies and select only a relatively small number of phases with high frequencies. Existing methods proposed in Refs. 12 and 16 also shows that incomplete matrix with low frequencies yield good reconstruction accuracy. (2) Moreover, the asymmetricity of the TITLE is more obvious than those of the Zhang method and the Pineda method. The asymmetricity is preferable because a K-space data is Hermitian symmetric in theory and symmetric phases are redundant to a great extent.

Figure 7 shows how the selected phases of the proposed TITLE method are adaptive to the contents of different slices. In the left column of Figure 7, most selected phases correspond to low frequencies neighboring to the pre-selection phases. From the left column to the right column, as the energy in low frequency decreases, more phases distant to the center of the pre-selection phases are chosen. The following trend can be seen. For slices with very large energy in low-frequency, there is very large possibility that phases with low frequency are selected. For slices with less energy in low-frequency, there is a large possibility that the TITLE algorithm switches to high-frequency exploration.

## 4 DISCUSSION AND CONCLUSION

We have presented a transformer-based sequential phase-encode selection method, called TITLE, for the purpose of actively generating a fast MR scan trajectory specific to each slice. The experimental results on the fastMRI dataset demonstrated that the TITLE method is not only much more efficient than the Pineda method but also yields better reconstruction performance in terms of SSIM, PSNR, and NSME no matter reconstruction is conducted by the advanced UNet or the simplest IFT.

The success of TITLE comes from introducing a phase token to a light-weight transformer network, employing fast Inverse Fourier Transform for computing reward of selecting a phase, using the transformer to estimate Q-values (a Q-value reflects the importance of a phase), integrating the transformer-based Q-network into the framework of deep reinforcement learning.

Generally, to achieve better or comparable accuracy for the same task, a transformer-based network requires much more parameters and computational cost than a state-of-the-art CNN. Moreover, a transformer-based model should be pre-trained on an additional large dataset. It is interesting that the proposed transformer-based method with smaller number of parameters and without pre-training on an additional large-scale dataset can outperform the CNN-based deep reinforcement learning method of the Pineda.

In the future, we plan to extend the proposed method from the Cartesian trajectory to the Radial and other types of trajectories. In addition, the proposed method will be applied on multi-coil data and extended for generating 3D scan trajectory.

## ACKNOWLEDGEMENT

We would like to thank Dr. Jinghua Wang for calculating the maximum of allowable phase selection and correcting typos of the manuscript.